\documentclass[]{spie}  

\usepackage{amsmath,amsfonts,amssymb}
\usepackage{graphicx}
\usepackage[colorlinks=true, allcolors=blue]{hyperref}
\usepackage{multirow,diagbox,floatrow,subfig,tabularx,placeins}
\newfloatcommand{capbtabbox}{table}[][\FBwidth]

\title{Generalization ability of region proposal networks for multispectral person detection}

\author[a,b]{Kevin Fritz}
\author[a]{Daniel K{\"o}nig}
\author[b]{Ulrich Klauck}
\author[a]{Michael Teutsch}
\affil[a]{Hensoldt Optronics GmbH, Carl-Zeiss-Str. 22, 73447 Oberkochen, Germany}
\affil[b]{Aalen University of Applied Sciences, Beethovenstr. 1, 73430 Aalen, Germany}

\authorinfo{Author email addresses: \{kevin.fritz, daniel.koenig, michael.teutsch\}@hensoldt.net, ulrich.klauck@hs-aalen.de}

\begin{document} 
\maketitle

\begin{abstract}
Multispectral person detection aims at automatically localizing humans in images that consist of multiple spectral bands. Usually, the visual-optical (VIS) and the thermal infrared (IR) spectra are combined to achieve higher robustness for person detection especially in insufficiently illuminated scenes. This paper focuses on analyzing existing detection approaches for their generalization ability. Generalization is a key feature for machine learning based detection algorithms that are supposed to perform well across different datasets. Inspired by recent literature regarding person detection in the VIS spectrum, we perform a cross-validation study to empirically determine the most promising dataset to train a well-generalizing detector. Therefore, we pick one reference Deep Convolutional Neural Network (DCNN) architecture as well as three different multispectral datasets. The Region Proposal Network (RPN) that was originally introduced for object detection within the popular Faster \mbox{R-CNN} is chosen as a reference DCNN. The reason for this choice is that a stand-alone RPN is able to serve as a competitive detector for two-class problems such as person detection. Furthermore, all current state-of-the-art approaches initially apply an RPN followed by individual classifiers. The three considered datasets are the KAIST Multispectral Pedestrian Benchmark including recently published improved annotations for training and testing, the Tokyo Multi-spectral Semantic Segmentation dataset, and the OSU Color-Thermal dataset including just recently released annotations. The experimental results show that the KAIST Multispectral Pedestrian Benchmark with its improved annotations provides the best basis to train a DCNN with good generalization ability compared to the other two multispectral datasets. On average, this detection model achieves a log-average Miss Rate (MR) of 29.74\,\% evaluated on the reasonable test subsets of the three analyzed datasets.
\end{abstract}

\keywords{Human Detection, Visual Surveillance, Multispectral Images, Deep Learning, DCNNs}

\section{INTRODUCTION}
\label{sec:intro}

Person detection is a widely discussed and analyzed field of research in computer vision~\cite{Dollar2012,Benenson2014}. There are many camera-based safety and security applications such as driver assistance, surveillance, or search and rescue. Current state-of-the-art approaches are well-performing and well-optimized but still prone to produce false positive (FP) and false negative (FN) detections due to occlusions, low resolution, weak illumination, or image perturbations such as noise and blur~\cite{Zhang2016a}. This is even the case for popular detection approaches that utilize deep learning techniques such as Deep Convolutional Neural Networks (DCNNs)~\cite{Hosang2015,Zhang2017,Brazil2017}. Most authors consider visual-optical (VIS) images only and are thus dependent on appropriate illumination conditions. Infrared (IR) images, however, are a promising source of data for improvement since persons usually have a prominent signature in the thermal IR spectrum~\cite{Teutsch2014}. In recent years, several authors demonstrated that the VIS and the thermal IR spectra provide strong synergies. Hence, for person detection multispectral fusion DCNNs are able to clearly outperform pure VIS based especially under difficult illumination conditions~\cite{Wagner2016,Liu2016,Koenig2017a,Li2018}.

In this paper, we discuss the problem of multispectral person detection. We aim at automatically localizing humans in images that consist of multiple spectral bands. These bands are assumed to be spatially aligned, i.e. at each pixel location we expect the same scene content across all spectra. Usually, the VIS and the thermal IR spectra are combined to achieve higher robustness for person detection especially in insufficiently illuminated scenes. The focus of this work lies on analyzing existing approaches for their generalization ability. Generalization is a key feature for detectors that are supposed to perform well across different datasets. While there are detector models that are trained for one camera or one dataset and thus perform well with this specific data only, the performance of well-generalizing models does not degrade significantly when applied to similar data originating from other cameras. There are many benefits given by well-generalizing models: (1) overfitting during training is implicitly avoided, (2) the model can be re-used for other cameras and datasets without the necessity of annotating new data for model adaptation, and (3) more training data can be acquired by collecting training samples from multiple cameras and datasets. Inspired from recent literature in the field of VIS person detection, we pick one reference DCNN architecture as well as three different datasets for multispectral person detection and perform a cross-validation study to empirically determine the most promising dataset to train a well-generalizing detector model. The well-known Region Proposal Network (RPN) architecture that was originally introduced for object detection within the popular Faster Regions with CNN features (\mbox{R-CNN}) architecture is chosen as a reference model. The reason for this choice is that a stand-alone RPN is already able to serve as a competitive detector for two-class problems such as person detection~\cite{Zhang2016b}. Furthermore, all current state-of-the-art approaches initially apply an RPN followed by individual classifiers such as Fast \mbox{R-CNN}~\cite{Zhang2017}, Boosted Forest (BF)~\cite{Zhang2016b}, or Binary Classification Network (BCN)~\cite{Brazil2017}. The three considered datasets are the KAIST Multispectral Pedestrian Benchmark~\cite{Hwang2015} including recently published improved annotations for training~\cite{Li2018} and testing~\cite{Liu2017}, the Tokyo Multi-spectral Semantic Segmentation dataset~\cite{Ha2017}, and the OSU Color-Thermal dataset~\cite{Davis2005} including just recently released annotations~\cite{Biswas2017}. In addition, several pure VIS and IR datasets such as the Caltech Pedestrian Detection Benchmark~\cite{Dollar2012}, CityPersons~\cite{Zhang2017} (both VIS), and \mbox{CVC-09} FIR Sequence Pedestrian Dataset~\cite{Socarras2013} are analyzed for their impact on the multispectral fusion during domain adaptation. Domain adaptation is necessary since we use a \mbox{VGG-16} network~\cite{Simonyan2015} as DCNN backbone that was pre-trained on ImageNet classification~\cite{Russakovsky2015}. For preparing the multispectral fusion, one DCNN for the VIS and one for the IR spectrum are trained individually and first need to be domain adapted from the task of image classification to person detection. Then, we follow prior work taken from the literature\cite{Koenig2017a,Koenig2017b} performing a halfway spectral fusion in order to ensure having a strong multispectral DCNN reference model for our experiments.

The remainder of this paper is organized as follows: related work is reviewed in Section~\ref{sec:relwork}, the three utilized multispectral datasets are presented in Section~\ref{sec:data}, experimental results are discussed in Section~\ref{sec:experiments}, and conclusions are given in Section~\ref{sec:conclu}.

\section{RELATED WORK}
\label{sec:relwork} 

\textbf{Multispectral person detection:} Some of the first publications on multispectral person detection are provided by Davis and Sharma~\cite{Davis2005,Davis2007} with presenting and exploiting the OSU Color-Thermal Database. Regions of interest (proposals) are generated via background subtraction and evaluated using contour based fusion. Other approaches for further evaluation of the foreground objects within this dataset were published by Leykin et al.~\cite{Leykin2007} evaluating periodic gait analysis, Y{\"u}r{\"u}k~\cite{Yuruk2008} using active contours, or \mbox{San-Biagio} et al.~\cite{SanBiagio2012} utilizing Riemannian manifolds. Then, multispectral person detection gained attention again with the introduction of the KAIST Multispectral Pedestrian Benchmark~\cite{Hwang2015}. Inspired by Doll{\'{a}}r et al.~\cite{Dollar2014}, a combination of Aggregated Channel Features (ACF) and BF is used to detect pedestrians in images with four channels: Red (R), Green (G), Blue (B), and Thermal (T). With its large extent and since a moving camera is used in an automotive setup, this dataset motivated many authors to propose new multispectral fusion approaches that are based on novel deep learning techniques. Wagner et al.~\cite{Wagner2016} generate proposals using ACF and BF, and classify them with a multispectral fusion DCNN. Liu et al.~\cite{Liu2016} adopt the Faster \mbox{R-CNN} architecture and identify the most promising layer or stage of VIS and IR information fusion within the DCNN. Choi et al.~\cite{Choi2016} apply VIS and IR RPNs individually and classify the generated proposals by evaluating fused deep features via Support Vector Regression (SVR). End-to-end training is made possible with a later published extension by Park et al.~\cite{Park2018}. Two of the most recent publications in the field of multispectral person detection that significantly improved the state-of-the-art adopted and adapted highly promising approaches taken from VIS person detection: while K{\"o}nig et al.~\cite{Koenig2017a} extended the \mbox{RPN+BF} approach~\cite{Zhang2016b}, Li et al.~\cite{Li2018} extended the Simultaneous Detection and Segmentation \mbox{R-CNN} (SDS-RCNN) architecture~\cite{Brazil2017}. Finally, the application of illumination-aware weighting functions for a late fusion of individual daytime and nighttime DCNNs~\cite{Guan2019} or individual color and thermal DCNNs~\cite{Li2019} has proven itself as promising.

\noindent \textbf{Generalization ability:} Generalization in computer vision and machine learning is the ability of image and video processing algorithms such as object detectors or classifiers to perform well not only on the dataset they are trained on but also on previously unseen data acquired by different cameras or under different environmental conditions such as daytime vs. nighttime. Usually, such datasets underlie different image quality perturbations such as specific sensor noise, contrast, or blur that can severely affect the processing quality~\cite{Geirhos2018}. Detector or classifier models exclusively and extensively trained on this data are prone to overfit~\cite{Bishop2006} or getting biased~\cite{Torralba2011}. In contrast to explicit methods that aim at reducing or avoiding such effects during model training~\cite{Srivastava2014,Ioffe2015,Neyshabur2017,Wu2018} or application~\cite{Ulen2014}, we are inspired by Torralba and Efros~\cite{Torralba2011} and Zhang et al.~\cite{Zhang2017} and compare models trained on different datasets in a cross-validation manner. In addition, different datasets for domain adaptation and transfer learning~\cite{Wagner2016,Koenig2017a,Koenig2017b} are exploited that best support the multispectral fusion.

\section{DATASETS}
\label{sec:data}

The six public datasets for person detection that are used in this work are described in more detail within this section. There are two pure VIS datasets with the Caltech Pedestrian Detection Benchmark and the CityPersons dataset, one pure thermal IR dataset with the \mbox{CVC-09} FIR Sequence Pedestrian Dataset, and three multispectral datasets with the KAIST Multispectral Pedestrian Benchmark, the OSU Color-Thermal dataset, and Tokyo Multi-spectral Semantic Segmentation. Figure~\ref{fig:datasets} shows an overview of the considered datasets with (1) their partitioning in training and testing subdatasets, (2) the extent of each subdataset regarding number of images and number of annotated persons represented by bounding boxes, and (3) the person height distribution depicted as histogram.

\subsection{Caltech Pedestrian Detection Benchmark}

\begin{figure}[ht]
\ffigbox{
  \begin{subfloatrow}
    \capbtabbox{
\begin{tabular}{|l|l|l|}
\hline
\textbf{Caltech}~\cite{Dollar2012} & Boxes & Images (Skip)\\
\hline
\hline
\textit{test-Reason./-new} & 1,076 / 912 & 4,024 (30) \\                   
\hline
\textit{test-All/-new} & 2,590 / 2,323 & 4,024 (30) \\         
\hline
\textit{train/-new} & 1,631 / 1,619 & 4,250 (30) \\   
\hline
\textit{train10x} & 16,376 & 42,782 (3) \\
   \hline   
    \end{tabular}
}{
  \caption{Caltech subdatasets and their extent.}
	\label{table:caltech}
}
\ffigbox{
  \includegraphics[height=2.25cm]{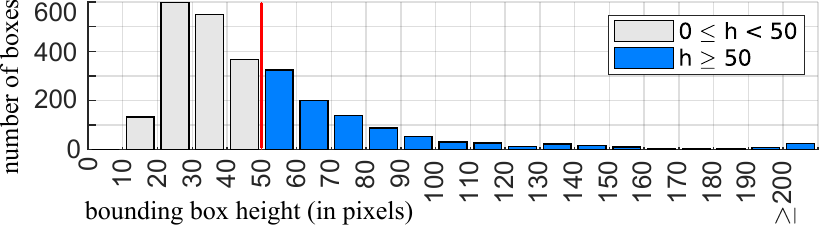}
}{
  \caption{Person height histogram for Caltech \textit{test-All-new}.}
	\label{fig:caltech_hist}
}
  \end{subfloatrow}
  \renewlengthtocommand\settowidth\Mylen{\subfloatrowsep}\vskip\Mylen
  \begin{subfloatrow}
    \capbtabbox{
\begin{tabular}{|l|l|l|}
\hline
\multirow{2}{*}{\textbf{CityPersons}~\cite{Zhang2017} \quad\qquad}  & \multirow{2}{*}{Boxes \quad} & \multirow{2}{*}{Images (Skip)}\\
&&\\
\hline
\hline
\textit{val-Reasonable} & 1,579 & 500 (1) \\                   
\hline
\textit{val-All} & 1,954 & 500 (1) \\         
\hline
\textit{train} & 7,891 & 2,975 (1) \\   
   \hline   
    \end{tabular}
}{
  \caption{CityPersons subdatasets and their extent.}
	\label{table:citypersons}
}
\ffigbox{
  \includegraphics[height=2.25cm]{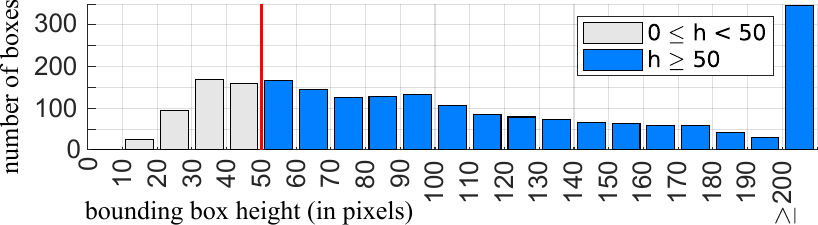}
}{
  \caption{Person height histogram for CityPersons \textit{val-All}.}
	\label{fig:citypersons_hist}
}
  \end{subfloatrow}
	\renewlengthtocommand\settowidth\Mylen{\subfloatrowsep}\vskip\Mylen
  \begin{subfloatrow}
    \capbtabbox{
\begin{tabular}{|l|l|l|}
\hline
{\textbf{CVC-09}~\cite{Socarras2013} \qquad\qquad\;\;} & {Boxes \quad} & Images (Skip)\\
\hline
\hline
\textit{test-Reasonable} & 1,018 & 432 (20) \\                   
\hline
\textit{test-All} & 1,052 & 432 (20) \\         
\hline
\textit{train} & 1,482 & 420 (20) \\   
\hline
\textit{train10x} & 15,058 & 4,209 (2) \\ 
   \hline   
    \end{tabular}
}{
  \caption{CVC-09 subdatasets and their extent.}
	\label{table:cvc}
}
\ffigbox{
  \includegraphics[height=2.25cm]{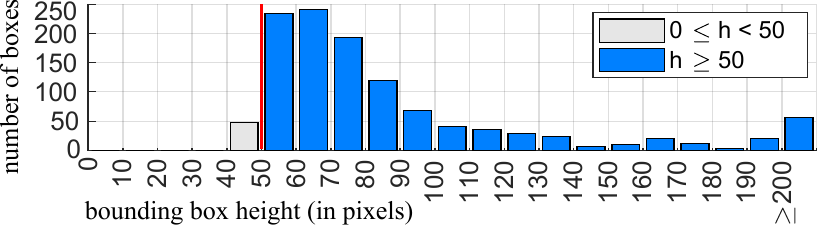}
}{
  \caption{Person height histogram for CVC-09 \textit{test-All}.}
	\label{fig:cvc_hist}
}
  \end{subfloatrow}
	\renewlengthtocommand\settowidth\Mylen{\subfloatrowsep}\vskip\Mylen
  \begin{subfloatrow}
    \capbtabbox{
\begin{tabular}{|l|l|l|}
\hline
\textbf{KAIST}~\cite{Hwang2015} & Boxes & Images (Skip)\\
\hline
\hline
\textit{test-Reason./-new} & 1,831 / 2,039 & 2,252 (20) \\                   
\hline
\textit{test-All/-new} & 2,245 / 3,390 & 2,252 (20) \\                   
\hline
\textit{train} & 1,748 & 2,500 (20) \\   
\hline
\textit{train10x} & 17,419 & 25,086 (2) \\  
\hline
\textit{train-new} & 16,509 & 7,601 (2) \\  
   \hline   
    \end{tabular}
}{
  \caption{KAIST Multispectral subdatasets and their extent.}
	\label{table:kaist}
}
\ffigbox{
  \includegraphics[height=2.7cm]{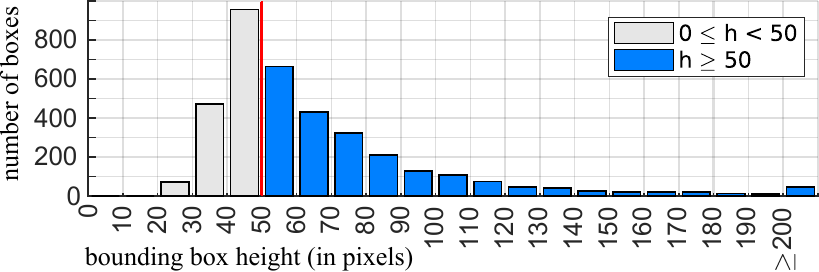}
}{
  \caption{Person height histogram for KAIST \textit{test-new-All}.}
	\label{fig:kaist_hist}
}
  \end{subfloatrow}
	\renewlengthtocommand\settowidth\Mylen{\subfloatrowsep}\vskip\Mylen
  \begin{subfloatrow}
    \capbtabbox{%
\begin{tabular}{|l|l|l|}
\hline
\textbf{OSU Color-Thermal}~\cite{Davis2005} & Boxes & Images (Skip)\\
\hline
\hline
\textit{scale2x-test-Reasonable} & 234 & 258 (20) \\                   
\hline
\textit{scale2x-test-All} & 234 & 258 (20) \\         
\hline
\textit{scale2x-train} & 527 & 167 (20) \\   
\hline
\textit{scale2x-train10x} & 5,315 & 1,667 (2) \\ 
   \hline   
    \end{tabular}
}{
  \caption{OSU Color-Thermal subdatasets and their extent.}
	\label{table:osu}
}
\ffigbox{
  \includegraphics[height=2.25cm]{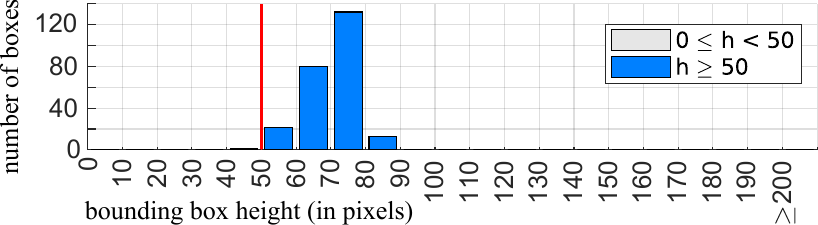}
}{
  \caption{Person height histogram for OSU \textit{test-All}.}
	\label{fig:osu_hist}
}
  \end{subfloatrow}
	\renewlengthtocommand\settowidth\Mylen{\subfloatrowsep}\vskip\Mylen
  \begin{subfloatrow}
    \capbtabbox{
\begin{tabular}{|l|l|l|}
\hline
\multirow{2}{*}{\textbf{Tokyo Segmentation}~\cite{Ha2017}} & \multirow{2}{*}{Boxes} & \multirow{2}{*}{Images (Skip)}\\
&&\\
\hline
\hline
\textit{testval-Reasonable} & 880 & 785 (1) \\                   
\hline
\textit{testval-All} & 1,561 & 785 (1) \\         
\hline
\textit{train} & 1,093 & 784 (1) \\   
   \hline   
    \end{tabular}
}{
  \caption{Tokyo Segmentation subdatasets and their extent.}
	\label{table:tokyo}
}
\ffigbox{
  \includegraphics[height=2.25cm]{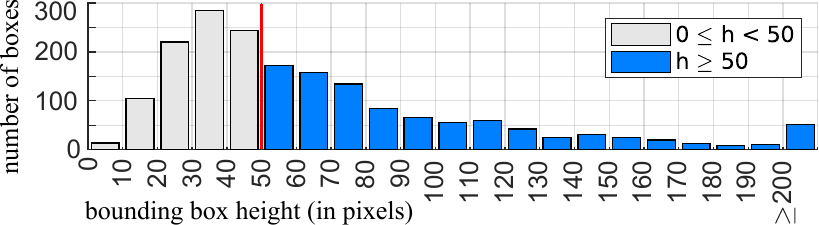}
}{
  \caption{Person height histogram for Tokyo \textit{testval-All}.}
	\label{fig:tokyo_hist}
}
  \end{subfloatrow}
}{\caption{Datasets used for experiments and their properties such as extent or annotations' height. The red line in each histogram shows the separation between reasonable (blue bars) and all (grey and blue bars) test data.}
  \label{fig:datasets}}
\end{figure}

The Caltech Pedestrian Detection Benchmark~\cite{Dollar2012} is one of the most widely utilized public datasets for tackling the problem of person detection in computer vision. The camera is mounted on a moving vehicle to cover a large variety of different scenes within an automotive setup. The original \textit{train} and \textit{test} subdatasets as proposed by Doll{\'a}r et al.~\cite{Dollar2012} comprise \mbox{1,631} and \mbox{2,590} annotated persons, respectively, as shown in Table~\ref{table:caltech}. To avoid overfitting due to highly similar scene background, only every 30th image in the image sequence is picked for training and testing. This parameter is called \emph{skip}. In addition, the differentiation between \textit{Reasonable} and \textit{All} data was introduced: annotations with a bounding box height of at least 50 pixels and no visible occlusion are considered \textit{Reasonable} while boxes of 20 pixels or more in height and at least 65\,\% of non-occluded visible area are denoted as \textit{All}. In this way, heavily occluded persons and the large number of small persons within the dataset that can hardly be detected automatically do not bias the evaluation. Since the training data seems to be not sufficient to train DCNNs, Hosang et al.~\cite{Hosang2015} proposed to reduce the skip to 3 in order to collect ten times more training samples. This subdataset is called \textit{train10x}. However, the original annotation data did not contain any information about occlusions. Instead, Zhang et al.~\cite{Zhang2017} revised the annotation data as it was impaired by imprecise interpolation between frames~\cite{Zhang2016a} and added information about occlusions. The new annotations use skip 30 and are denoted by \textit{new} in Fig.~\ref{table:caltech}. The number of annotations is slightly reduced, which is the result of the introduction of ignore regions that cover image areas with no unambiguous assignment of annotations. Bounding boxes of the \textit{new} subdatasets fit closer to a person compared to the original annotation data. The distribution of bounding box height is presented in the histogram shown in Fig.~\ref{fig:caltech_hist}. The red line visually separates the \textit{Reasonable} and \textit{All} parts of the \textit{test-All-new} subdataset. We can recognize the large number of rather small persons with a height of 50 pixels or less within the dataset.

\FloatBarrier

\subsection{CityPersons}

The CityPersons dataset~\cite{Zhang2017} is a re-labeling of the popular Cityscapes dataset~\cite{Cordts2016} that was introduced to tackle the challenge of semantic segmentation in inner cities for automotive applications. The annotation protocol is arranged in a way that bounding boxes fit closely to the labeled persons within the images. Hence, there is only very few background inside the image regions that contain persons. The assumption is that in this way the background less influences the training process and thus the generalization ability of a model trained with CityPersons is expected to increase. This assumption is validated for a Faster \mbox{R-CNN} based person detector that generalizes well across different person detection benchmark datasets such as Caltech~\cite{Dollar2012} or KITTI~\cite{Geiger2012}. In addition, it is demonstrated that pre-training with CityPersons before performing the regular training for Caltech or KITTI can further improve the resulting models. This motivates us to consider CityPersons in our experiments, too, for pre-training and domain adaptation. Since the test subdataset is not publicly available, we use the validation data that is denoted by \textit{val}. As seen in Fig.~\ref{table:citypersons}, the \textit{train} and \textit{val} subdatasets contain \mbox{7,891} and \mbox{1,954} annotated persons, respectively. They are adopted as proposed by Zhang et al.~\cite{Zhang2017}. The distribution of the bounding box heights is visualized in Fig.~\ref{fig:citypersons_hist}. Due to the high image resolution of $2,048\times1,024$ pixels, a high peak occurs for persons with a height of 200 pixels or more. Compared to the other considered datasets, the person heights within the CityPersons dataset are more uniformly distributed and the resolution is much higher: while the KAIST dataset has a resolution of $640\times512$ pixels, the four remaining datasets have a resolution of $640\times480$ pixels. Another interesting aspect of the dataset is that an additional bounding box is provided for each annotated person that contains the visible part of the person. This box is generated automatically by using the semantic segmentation labels and fits very closely to the labeled person. By comparing the visible and the manually annotated bounding box, the percentage of occlusion can be calculated precisely. We only consider persons for training that are at least 70\,\% visible in our experiments.

\subsection{CVC-09 FIR Sequence Pedestrian Dataset}

The \mbox{CVC-09} FIR Sequence Pedestrian Dataset~\cite{Socarras2013} is an automotive dataset acquired in the thermal IR spectrum. The annotations do not contain any information about ignore regions or occluded persons. Furthermore, there is a quite large amount of missing labels~\cite{Koenig2017b}. However, due to the vast size of the dataset, it is helpful for pre-training the individual IR RPN and thus performing the domain adaptation with real thermal IR data instead of synthesized data~\cite{Koenig2017a}. The distribution of \textit{train} and \textit{test} subdatasets shown in Fig.~\ref{table:cvc} is taken from the original paper~\cite{Socarras2013}. By using the reduced image skip of 2 instead of 20 (\textit{train10x}), the number of training samples can be increased from \mbox{1,482} to \mbox{15,058}. When having a closer look at the red line in the related histogram of the \textit{test-all} subdataset in Fig.~\ref{fig:cvc_hist}, we can see that the dataset mostly contains persons of height larger than 50 pixels.

\subsection{KAIST Multispectral Pedestrian Benchmark}

The KAIST Multispectral Pedestrian Detection Benchmark~\cite{Hwang2015} is currently the most popular benchmark dataset for multispectral person detection and combines the VIS and the thermal IR spectra. Just like the already mentioned datasets, it is an automotive dataset, too. The problem of having different viewing angles of the VIS and the IR camera is avoided by using a beam splitter. Each image consists of four channels: red (R), green (G), blue (B), and thermal IR (T). The original \textit{train} and \textit{test} subdatasets comprise of \mbox{1,748} and \mbox{2,245} labeled persons, respectively. Since the original annotations contain some mistakes such as wrong or missing labels~\cite{Koenig2017a}, the \textit{test} subdataset was corrected by Lui et al.~\cite{Liu2017} while the \textit{train} set was partially revised by Li et al.~\cite{Li2018}. The distribution of \textit{train} and \textit{test} subdatasets as seen in Fig.~\ref{table:kaist} is adopted from the original paper~\cite{Hwang2015} and the revised annotations are indicated with the label \textit{new}. It is notable that the \textit{train-new} subdataset contains annotations in a much higher density, i.e. the ratio between number of annotations and number of images, compared to the other \textit{train} datasets. The reason is that Li et al.~\cite{Li2018} removed all images that do not contain any persons. In addition, all images were removed that include highly occluded persons and very small persons with a height of less than 50 pixels. The impact of the corrected missing labels within the test dataset becomes apparent when comparing the \textit{test-Reasonable} and the \textit{test-Reasonable-new} subdatasets: about 10\,\% more labels appear in \textit{test-Reasonable-new}. The distribution of the labeled persons' height is plotted as a histogram in Fig.~\ref{fig:kaist_hist}. The red line again separates the blue annotations contained in both \textit{test-Reasonable-new} and \textit{test-All-new} and the gray annotations that are part of the \textit{test-All-new} dataset only. Compared to the original annotation's histogram~\cite{Koenig2017b}, the proportion of persons with a height between 40 and 60 pixels increased remarkably.

\subsection{OSU Color-Thermal}

The OSU Color-Thermal dataset~\cite{Davis2005} is the only dataset considered here that is not originating from automotive but from surveillance applications. Just like the KAIST dataset, it consists of three color VIS channels and one thermal IR channel. The original resolution is $320\times240$ pixels but we upscaled the images to a resolution of $640\times480$ pixels (denoted by \textit{scale2x}) in order to align the dataset with the others. This is of specific interest when determining the \textit{Reasonable} subdatasets, in which a person must have a height of 50 pixels or more inside the image. Although the dataset was published several years ago back in 2005 already, annotations for the person detection task were not available until 2017~\cite{Biswas2017}. There are six image sequences in total acquired by a stationary sensor mounted on a building. There is no recommendation for splitting the data in train and test subdatasets. Since only two different scenes are observed, however, we choose sequences 1-3 that show the first scene for training and sequences 4-6 that show the second scene for testing. The \textit{train} and \textit{test} subdatasets as shown in Fig.~\ref{table:osu} comprise \mbox{527} and \mbox{234} annotated persons only. To increase the number of training samples, the image skip is reduced from 20 to 2 (\textit{train10x}). In this way, we can get \mbox{5,315} samples. But since the camera is stationary, we expect a high risk of overfitting as only one scene is observed and hence the background is constant within the training data. Furthermore, there are synchronization issues due to occasionally occurring frame drops in the VIS or the IR channel. The distribution of the bounding box heights is plotted in Fig.~\ref{fig:osu_hist}. Due to the strong limitations of this dataset with a stationary sensor and only two different scenes, only persons between 50 and 90 pixels in height appear.

\subsection{Tokyo Multi-spectral Semantic Segmentation}
\label{subsec:tokyodata}

Tokyo Multi-spectral Semantic Segmentation dataset~\cite{Ha2017} is another automotive dataset. Just like the Cityscapes dataset, the challenge of semantic segmentation is addressed. However, since it is a multispectral dataset, again three color VIS channels and one thermal IR channel are provided. About half of the \mbox{1,569} images are acquired by night. We generate bounding box annotations from the pixelwise semantic labels by creating one bounding box for each set of connected pixels that are labeled as person. The bounding box borders are determined by searching the leftmost, rightmost, upmost, and bottommost pixel of the set. To avoid occluded persons in the data, we label bounding boxes with a width/height ratio lower than 0.2 and higher than 0.6 as ignore regions. We split the data in train and test subdatasets as proposed by Ha et al.~\cite{Ha2017}. However, we unite the validation and the test set to a larger set denoted \textit{testval}. In this way, we can split the data in 50\,\% for training and 50\,\% for testing. As seen in Fig.~\ref{table:tokyo}, we can generate \mbox{1,093} training and \mbox{1,561} test samples, respectively. To handle the rather small number of training samples, we use flipping for data augmentation to increase the number of samples during training. The distribution of the bounding box heights in Fig.~\ref{fig:tokyo_hist} shows that similar to Caltech there is a high percentage of small persons with a height of 50 pixels or less within the dataset.

\section{EXPERIMENTS}
\label{sec:experiments}

The experiments section is organized in three parts: we describe the experimental setup including the prior work we built on, the chosen DCNN architecture, and the training strategy in Section~\ref{subsec:expsetup}. The results presented in Section~\ref{subsec:expres} show the influence of various datasets for pre-training and fine-tuning applied to individual RPN models for the VIS and the IR spectra. In addition, the generalization ability of a multispectral fusion RPN across different datasets is evaluated. Finally, in Section~\ref{subsec:expdisc}, specific findings and observations of interest are discussed.

\subsection{Experimental Setup}
\label{subsec:expsetup}

Our experimental setup is built on the prior work of K{\"o}nig et al.~\cite{Koenig2017a,Koenig2017b}, in which the \mbox{RPN+BF} approach by Zhang et al.~\cite{Zhang2016b} is extended from pure VIS to multispectral images. Originally, the RPN is the first stage of the two-stage state-of-the-art object detection approach Faster \mbox{R-CNN}. The RPN is a proposal generating DCNN within the Faster \mbox{R-CNN} architecture that is followed by a classification DCNN called Fast \mbox{R-CNN}~\cite{Girshick2015}. With \mbox{RPN+BF}, it was demonstrated that a stand-alone RPN is able to outperform the Faster \mbox{R-CNN} approach in the domain of person detection. Since person detection is a two-class classification and localization problem, the RPN that usually separates objects and non-objects is able to tackle this challenge even without the classification stage of Faster \mbox{R-CNN}. The performance was even more improved by adding instead of a classification DCNN a rather simple classifier on top: the BF. Subsequently, however, it was shown that end-to-end trainable DCNNs tailored for person detection such as a slightly modified Faster \mbox{R-CNN}~\cite{Zhang2017} or an RPN in combination with a BCN~\cite{Brazil2017} are able to at least perform similar or better. The key finding is that each recent approach, which improved the state-of-the-art, was based on the initial application of an RPN. This motivates us to assume that a well-generalizing RPN is crucial for a well-generalizing person detector and thus specifically consider an RPN for our experiments. The multispectral fusion RPN by K{\"o}nig et al.~\cite{Koenig2017a} is inspired by prior work~\cite{Liu2016,Wagner2016}, in which a VIS DCNN and an IR DCNN are pre-trained individually with person detection datasets for domain adaptation in both spectra independently and then fused after a certain layer to create a multispectral DCNN.

Our RPN originates from the MATLAB code of the \mbox{RPN+BF} approach that was extended to act as a multispectral fusion RPN~\cite{Koenig2017a}. The implementation is based on the Caffe deep learning framework~\cite{Jia2014} and uses \mbox{VGG-16}~\cite{Simonyan2015} as DCNN backbone. Since VGG-16 network expects three planes per input image (RGB), we clone the single plane of each thermal IR image in order to create a three plane IR image as input for the IR RPN. As proposed and applied by several authors~\cite{Zhang2016a,Zhang2016b,Zhang2017}, the bounding box aspect ratio between width and height is fixed to 0.41 for each anchor within the RPN. If bounding boxes intersect each other with an overlap of at least 0.7 regarding the Intersection over Union (IoU) criterion, non-maximum suppression is used to keep the one with the highest confidence score while rejecting the others. The fusion RPN fuses the feature maps of the individual VIS RPN and IR RPN after the conv3 layer as suggested by several authors~\cite{Koenig2017a,Li2018}. The quantitative evaluation is based on the log-average Miss Rate (MR), which is the standard measure for pedestrian detection for many years now. It is a single number that averages the miss rates within an acceptable range of False Positives Per Image (FPPI). The sampling points for averaging are equidistantly distributed in the FPPI range that is logarithmic scaled. Since the miss rate indicates the rate of not detected persons (false negatives), a lower value represents a better person detector. Just like in most related literature, the evaluation is carried out for the reasonable testsets of each aforementioned dataset, i.e. persons with a bounding box height of at least 50 pixels and 35\,\% of occlusion at the maximum. Consequently, the reported MR is calculated for the reasonable testsets only.

\subsection{Results}
\label{subsec:expres}

At the beginning of the results subsection, we analyze different datasets or dataset combinations for pre-training and domain adaptation of the individual VIS and IR RPNs. For this experiment, we stay within the individual benchmarks and do not evaluate across datasets. Instead, we set up an RPN just for the VIS spectrum of the KAIST Multispectral Pedestrian Benchmark and the Tokyo Multi-spectral Semantic Segmentation dataset. Hence, this experiment consists of two individual parts: one for each dataset. The results are shown in Table~\ref{table:visrpn}. Different variants of pre-training with different datasets are tested. \mbox{Caltech-10x} denotes the \mbox{Caltech-\textit{train10x}} dataset as mentioned in Fig.~\ref{table:caltech}. Then, each pre-trained RPN is further trained with the VIS spectrum of the original multispectral training data: for the KAIST dataset, this is \mbox{KAIST-\textit{train-new}}, and for the Tokyo dataset, we use \mbox{Tokyo-\textit{train}}. In order to enrich the \mbox{Tokyo-\textit{train}} dataset, we introduce flipping for data augmentation as already mentioned in Section~\ref{subsec:tokyodata}. We skip the OSU Color-Thermal dataset for this experiment due to its strong biases. The best result for each dataset is highlighted in bold digits.

\begin{table}[!htb]
    \caption{VIS RPN training for the KAIST and the Tokyo dataset with different variants of pre-training. Each RPN is pre-trained with one of the listed dataset(s), then trained with KAIST-\emph{train-new} and Tokyo-\emph{train}, respectively, and finally evaluated on KAIST-\emph{test-Reasonable-new} and Tokyo-\emph{testval-Reasonable}, respectively.}
		\label{table:visrpn}
    \begin{minipage}{.5\linewidth}
      \centering
        \begin{tabular}{|c|c|}
				    \hline
				    \multicolumn{2}{|c|}{\textbf{KAIST VIS RPN}} \\
				    \hline
            Pre-Training Datasets & MR (\%) \\
						\hline
						\hline
						- & 42.48 \\
						\hline
						Caltech-10x & \textbf{38.15} \\
						\hline
						CityPersons & 41.31 \\
						\hline
						CityPersons + Caltech-new & 39.39 \\
						\hline
        \end{tabular}
    \end{minipage}%
    \begin{minipage}{.5\linewidth}
      \centering
        \begin{tabular}{|c|c|}
				    \hline
            \multicolumn{2}{|c|}{\textbf{Tokyo VIS RPN}} \\
				    \hline
            Pre-Training Datasets & MR (\%) \\
						\hline
						\hline
						- & 34.92 \\
						\hline
						Caltech-10x & 36.67 \\
						\hline
						CityPersons & \textbf{33.62} \\
						\hline
						CityPersons + Caltech-new & 34.04 \\
						\hline
        \end{tabular}
    \end{minipage} 
\end{table}

The results show that the MR seems not to be significantly influenced by the pre-training strategy. There is a remarkable decrease in MR for the KAIST dataset pre-trained with \mbox{Caltech-\textit{train10x}}. This might be conform with the `There's no data like more data' paradigm. However, we did not prove the significance of this MR decrease. Furthermore, this tendency cannot be confirmed with the Tokyo dataset that achieves its best result for pre-training with the \mbox{CityPersons-\textit{train}} dataset.

\begin{table}[!htb]
    \caption{IR RPN training for the KAIST and the Tokyo dataset with different variants of pre-training. Each RPN is pre-trained with one of the listed dataset(s), then trained with KAIST-\emph{train-new} and Tokyo-\emph{train}, respectively, and finally evaluated on KAIST-\emph{test-Reasonable-new} and Tokyo-\emph{testval-Reasonable}, respectively.}
		\label{table:irrpn}
    \begin{minipage}{.5\linewidth}
      \centering
        \begin{tabular}{|c|c|}
				    \hline
				    \multicolumn{2}{|c|}{\textbf{KAIST IR RPN}} \\
				    \hline
            Pre-Training Datasets & MR (\%) \\
						\hline
						\hline
						- & 29.49 \\
						\hline
						CVC-10x & \textbf{26.85} \\
						\hline
						CityPersonsR & 30.20 \\
						\hline
						CityPersonsR + CaltechR-new & 31.63 \\
						\hline
        \end{tabular}
    \end{minipage}%
    \begin{minipage}{.5\linewidth}
      \centering
        \begin{tabular}{|c|c|}
				    \hline
            \multicolumn{2}{|c|}{\textbf{Tokyo IR RPN}} \\
				    \hline
            Pre-Training Datasets & MR (\%) \\
						\hline
						\hline
						- & 10.07 \\
						\hline
						CVC-10x & 9.41 \\
						\hline
						CityPersonsR & \textbf{9.04} \\
						\hline
						CityPersonsR + CaltechR-new & 10.36 \\
						\hline
        \end{tabular}
    \end{minipage} 
\end{table}

The next experiment is essentially the same as the just mentioned one but now we consider the IR spectrum only. The related MRs can be seen in Table~\ref{table:irrpn}. One general finding is that the MR decreases and thus the detection results get better when using the IR instead of the VIS spectrum. Since both the KAIST and the Tokyo dataset contain night scenes with insufficient illumination for the VIS spectrum, this result is not really surprising. Due to the lack of IR datasets for pre-training, we not only consider the \mbox{CVC-09} dataset but also the red channel of the CityPersons and the Caltech datasets denoted by CityPersonsR and CaltechR, respectively, to somehow simulate the thermal IR channel. This approach is inspired by recent literature~\cite{Wagner2016,Koenig2017a}. Interestingly, we get similar results compared to the experiment in the VIS spectrum: while the KAIST dataset seems to benefit from the largest dataset \mbox{CVC-09}, the Tokyo dataset again performs best for pre-training with the CityPersons dataset. This is surprising because on the one hand the \mbox{CVC-09} dataset suffers from imprecise and missing annotations and on the other hand the CityPersons dataset is not an IR dataset. Again, however, on the basis of the presented results, the significance of the results cannot be proved here.

\begin{table}[!htb]
    \caption{Fusion RPN training for the KAIST and the Tokyo dataset with different variants of pre-training. Each VIS or IR RPN is pre-trained with one of the listed dataset(s) and trained with VIS/IR KAIST-\emph{train-new} and VIS/IR Tokyo-\emph{train}, respectively. Then, the fusion RPN is trained again with KAIST-\emph{train-new} and Tokyo-\emph{train}, respectively, and finally evaluated on KAIST-\emph{test-Reasonable-new} and Tokyo-\emph{testval-Reasonable}.}
		\label{table:fusrpn}
    \begin{minipage}{.5\linewidth}
      \centering
        \begin{tabular}{|c|c|c|}
				    \hline
				    \multicolumn{3}{|c|}{\textbf{KAIST Fusion RPN}} \\
				    \hline
            \multicolumn{2}{|c|}{Pre-Training Datasets} & \multirow{2}{*}{MR (\%)} \\
						\cline{1-2}
						VIS & IR & \\
						\hline
						\hline
						- & - & 21.97 \\
						\hline
						CityPersons & CityPersonsR & 22.35 \\
						\hline
						Caltech-10x & CVC-10x & 21.99 \\
						\hline
						CityPersons + & CityPersonsR + & \multirow{2}{*}{\textbf{21.46}} \\
						Caltech-new & CaltechR-new & \\
						\hline
						CityPersons + & \multirow{2}{*}{CVC-10x} & \multirow{2}{*}{22.42} \\
						Caltech-new & & \\
						\hline
        \end{tabular}
    \end{minipage}%
    \begin{minipage}{.5\linewidth}
      \centering
        \begin{tabular}{|c|c|c|}
				    \hline
				    \multicolumn{3}{|c|}{\textbf{Tokyo Fusion RPN}} \\
				    \hline
            \multicolumn{2}{|c|}{Pre-Training Datasets} & \multirow{2}{*}{MR (\%)} \\
						\cline{1-2}
						VIS & IR & \\
						\hline
						\hline
						- & - & 11.46 \\
						\hline
						CityPersons & CityPersonsR & 10.99 \\
						\hline
						Caltech-10x & CVC-10x & 12.49 \\
						\hline
						CityPersons + & CityPersonsR + & \multirow{2}{*}{\textbf{10.31}} \\
						Caltech-new & CaltechR-new & \\
						\hline
						CityPersons + & \multirow{2}{*}{CVC-10x} & \multirow{2}{*}{11.54} \\
						Caltech-new & & \\
						\hline
        \end{tabular}
    \end{minipage} 
\end{table}

The final dataset specific experiment is conducted for the fusion RPN. We pre-train the VIS RPN with the KAIST VIS dataset for the KAIST specific experiment in Table~\ref{table:fusrpn} and with the Tokyo VIS for the Tokyo specific experiment in Table~\ref{table:fusrpn}. The same pre-training strategy is applied analogously for the IR RPN. Then, we again evaluate different dataset combinations for pre-training the fusion RPN. Thereafter, the fusion RPN is trained with the multispectral \mbox{KAIST-\textit{train-new}} and \mbox{Tokyo-\textit{train}} datasets, respectively. The final evaluation is carried out on \mbox{KAIST-\textit{test-Reasonable-new}} and \mbox{Tokyo-\textit{testval-Reasonable}}, respectively. Table~\ref{table:fusrpn} shows the resulting MRs. Again, the results are very similar and the tendencies of the prior experiments cannot be confirmed in any way. One interesting finding is that the Tokyo IR channel seems to contain more helpful information for person detection compared to the multispectral data since the MRs consistently increase between Table~\ref{table:irrpn} and Table~\ref{table:fusrpn}.

As we still expect the \mbox{CityPersons-\textit{train}} and the \mbox{Caltech-\textit{train-new}} datasets to be the most promising ones for pre-training a well-generalizing VIS RPN model, and the \mbox{CVC-09-\textit{train10x}} to be the best dataset for pre-training the IR RPN model, we fix the pre-training strategy to this dataset combination for the following experiment. Fusion RPN pre-training is not applied. In addition, we also evaluate the OSU Color-Thermal dataset with this pre-training strategy and achieve a MR of 26.25\,\% for the \mbox{OSU-\textit{scale2x-test-Reasonable}} dataset trained with \mbox{OSU-\textit{scale2x-train10x}}.

\begin{table}
  \caption{Generalization of the multispectral fusion RPN across three multispectral datasets.}
  \label{tab:genab}
    \begin{center}
			\begin{tabular}{|c||c|c|c|}
			\hline
			\diagbox[width=38mm]{Test}{Train} & KAIST Multispectral & Tokyo Multi-spectral & OSU Color-Thermal \\
		  \hline
		  \hline
			KAIST Multispectral	& 22.42	& 71.04	& 91.35 	\\
			\hline
			Tokyo Multi-spectral & 34.97	& 11.54	& 78.93 	\\
			\hline
			OSU Color-Thermal	& 31.83	& 36.76	& 26.17 	\\
		  \hline
			\hline
		  Average	& \textbf{29.74}	& 39.78	& 65.48 \\
			\hline
			\multicolumn{1}{c|}{} & \multicolumn{3}{c|}{\textbf{MR (\%)}}  \\ 
			\cline{2-4}
			\end{tabular}
    \end{center}
\end{table}

In the remainder of this subsection, we analyze the performance of the multispectral fusion RPN models across datasets. Three fusion RPNs are trained in total: one for the KAIST Multispectral Pedestrian Benchmark, one for the Tokyo Multi-spectral Semantic Segmentation, and one for the OSU Color-Thermal dataset. Each model is pre-trained with the \mbox{CityPersons-\textit{train}} and the \mbox{Caltech-\textit{train-new}} datasets for the VIS RPN and the \mbox{CVC-09-\textit{train10x}} for the IR RPN. Then, the VIS and the IR RPNs are further trained separately using each dataset's training samples split in VIS and IR data. This means that for example the KAIST VIS RPN is trained with the VIS spectrum of the \mbox{KAIST-\textit{train10x}} dataset. Analogously, this training approach is applied to the IR spectrum. The fusion RPN is then trained with the multispectral training data of each dataset. Each of the resulting multispectral fusion RPNs is evaluated with the \mbox{KAIST-\textit{test-Reasonable-new}}, the \mbox{Tokyo-\textit{testval-Reasonable}}, and the \mbox{OSU-\textit{scale2x-test-Reasonable}}. The MRs for the individual datasets as well as the averaged MRs are presented in Table~\ref{tab:genab}. By far the best result is achieved by the KAIST RPN with an averaged MR of 29.74\,\%. The best individual performance is reached on the \mbox{KAIST-\textit{test-Reasonable-new}} with 22.42\,\%. However, the maximum gap between the best and the worst result is rather small with only 12.55\,\%. The Tokyo RPN provides the second best result with a MR of 39.78\,\% in average. However, we can see a huge gap between the \mbox{Tokyo-\textit{testval-Reasonable}} with 11.54\,\% and the \mbox{KAIST-\textit{test-Reasonable-new}} with 71.04\,\%. It seems that the training dataset is not sufficiently large to generalize well across datasets although the performance within the dataset is good. The OSU RPN performs worst by a large margin. The \mbox{OSU-\textit{scale2x-train10x}} has \mbox{5,315} training samples, which is much more compared to the \mbox{Tokyo-\textit{train}} dataset with \mbox{1,093} samples. However, the strong biases of the OSU Color-Thermal dataset with (1) a stationary camera only observing two different scenes, (2) the very low variance in scale of the persons present in the images, and (3) the nearly constant camera view angle prevent the resulting fusion RPN from generalizing well.

\begin{figure}[ht]
\centering
\includegraphics[width=\textwidth]{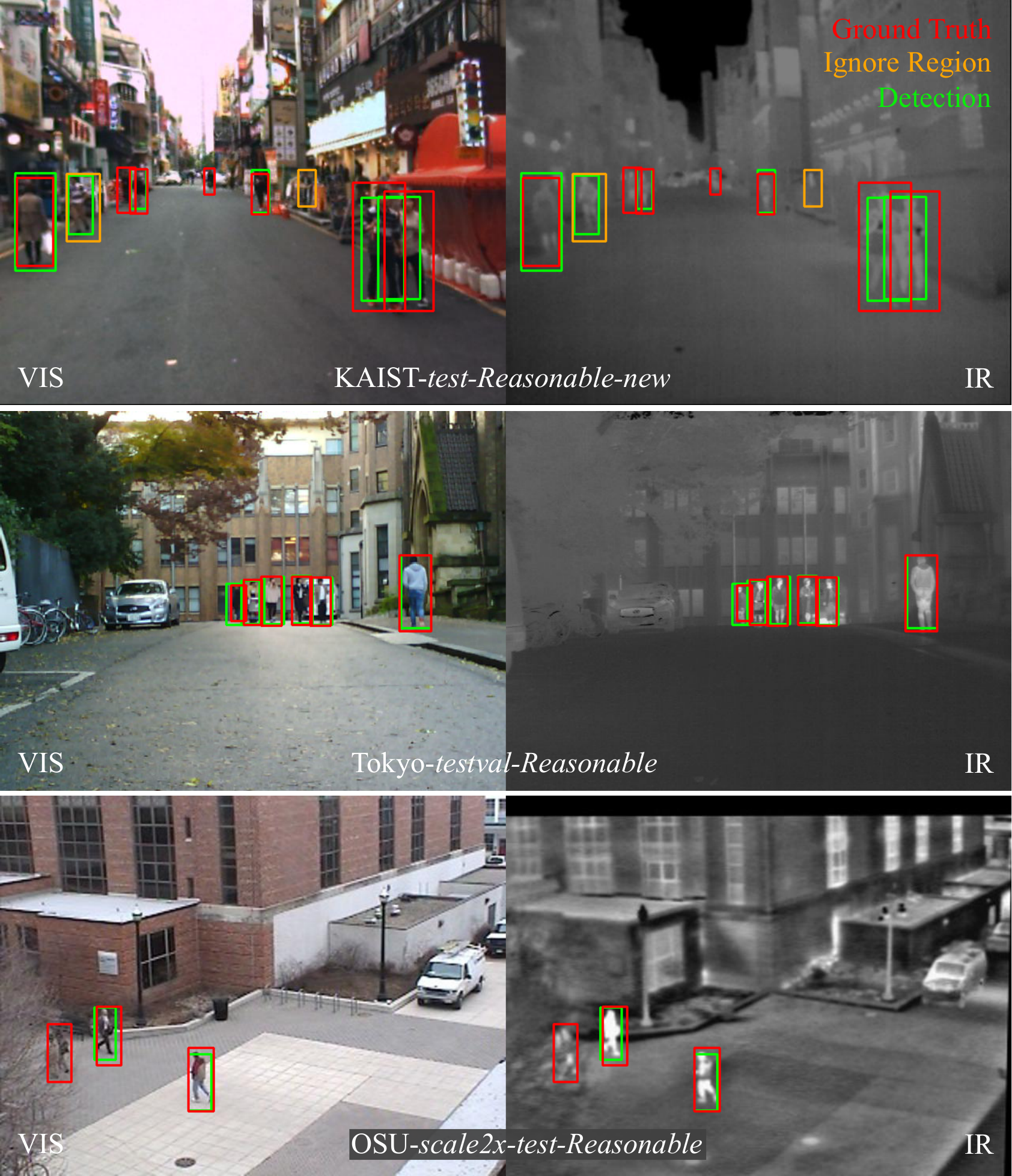}
  \caption{The best generalizing KAIST model according to Table~\ref{tab:genab} applied to the \mbox{KAIST-\textit{test-Reasonable-new}}, the \mbox{Tokyo-\textit{testval-Reasonable}}, and the \mbox{OSU-\textit{scale2x-test}} dataset.}
					\label{fig:examples}
\end{figure}

Some visual impressions of the results are given in Fig.~\ref{fig:examples}. We use the best generalizing model according to Table~\ref{tab:genab}, which is the fusion RPN pre-trained with \mbox{CityPersons-\textit{train}}, \mbox{Caltech-\textit{train-new}}, and \mbox{CVC-09-\textit{train10x}} and finally trained with the \mbox{KAIST-\textit{train-new}} dataset. The overall performance across the datasets is promising. The manually annotated ground truth is visualized in red color while the automatically determined detections are depicted in green color. The orange color indicates ignore regions, where the annotator could not gain sufficient certainty about the presence of one or more potentially highly occluded persons. Considering the example detections, we can see that the multispectral fusion RPN tends to produce bounding boxes fitting closer to the persons compared to prior work~\cite{Koenig2017a}. This may be the result of pre-training with the \mbox{CityPersons-\textit{train}} and the \mbox{Caltech-\textit{train-new}} datasets in the VIS spectrum since both datasets are labeled with closely fitting bounding boxes to minimize the influence of the background surrounding the persons during training.

\FloatBarrier

\subsection{Discussion}
\label{subsec:expdisc}

Further findings and interpretation of the results are presented and discussed in this subsection. 

\noindent \textbf{More accurate or just more training data?} In Section~\ref{sec:data}, we mentioned the revised annotations of the Caltech and the KAIST datasets, namely \mbox{Caltech-\textit{train-new}} and \mbox{KAIST-\textit{train-new}}. Missing annotations were added and bounding boxes that contain too much background were corrected to fit closer to the annotated persons for example. These revisions were published for the original training datasets only with skip 30 for Caltech and 20 for KAIST. Hosang et al.~\cite{Hosang2015}, however, showed that larger training datasets such as the \mbox{Caltech-\textit{train10x}} or the \mbox{KAIST-\textit{train10x}} are better suited for training DCNNs. Our results in Tables~\ref{table:visrpn}, \ref{table:irrpn}, and \ref{table:fusrpn} do not show any preference for either the \mbox{Caltech-\textit{train-new}} or the \mbox{Caltech-\textit{train10x}} datasets for pre-training. Further experiments that we do not include here even demonstrated that there is no benefit when using the improved annotations \mbox{KAIST-\textit{train-new}} instead of the error-prone \mbox{KAIST-\textit{train10x}} dataset. The large extend of the \mbox{KAIST-\textit{train10x}} dataset seems to compensate for the annotation mistakes.

\noindent \textbf{Is pre-training beneficial?} In the domain of multispectral person detection with the training datasets considered in this paper, pre-training with different training datasets than the target dataset did not show any significant improvement during the conducted experiments. The results presented in Tables~\ref{table:visrpn} and \ref{table:irrpn} appeared to be promising since some of the pre-training datasets such as \mbox{Caltech-\textit{train10x}} and the \mbox{CVC-09-\textit{train10x}} for KAIST and CityPersons for the Tokyo dataset reduced the MR remarkably. However, after the multispectral fusion training, no consistent and significant improvement of the MR can be proved anymore as shown in Table~\ref{table:fusrpn}. Instead, the pre-training datasets that generated the best results for the individual VIS and IR RPNs (Tables~\ref{table:visrpn} and \ref{table:irrpn}) do not clearly contribute to the training dataset combination that produced the best MRs for the fusion RPNs. This can be seen when comparing the performance of the pre-trained KAIST fusion RPN evaluated on \mbox{KAIST-\textit{test-Reasonable-new}} with a MR of 22.42\,\% reported in Table~\ref{tab:genab} with the not pre-trained KAIST fusion RPN with a MR of 21.97\,\% shown in Table~\ref{table:fusrpn}. Competitive MRs occur even when omitting pre-training for the fusion RPN. However, we do need intra-dataset pre-training for the individual VIS and IR RPNs: if we directly apply fusion RPN training without any pre-training for the KAIST VIS and IR RPN, we achieve a MR of 24.53\,\% for the fusion RPN. If we pre-train the VIS RPN with the KAIST VIS training dataset and the IR RPN with the KAIST IR dataset, the MR for the fusion RPN decreases to 21.97\,\% as reported in Table~\ref{table:fusrpn}.

\begin{figure}[ht]
\centering
\includegraphics[width=\textwidth]{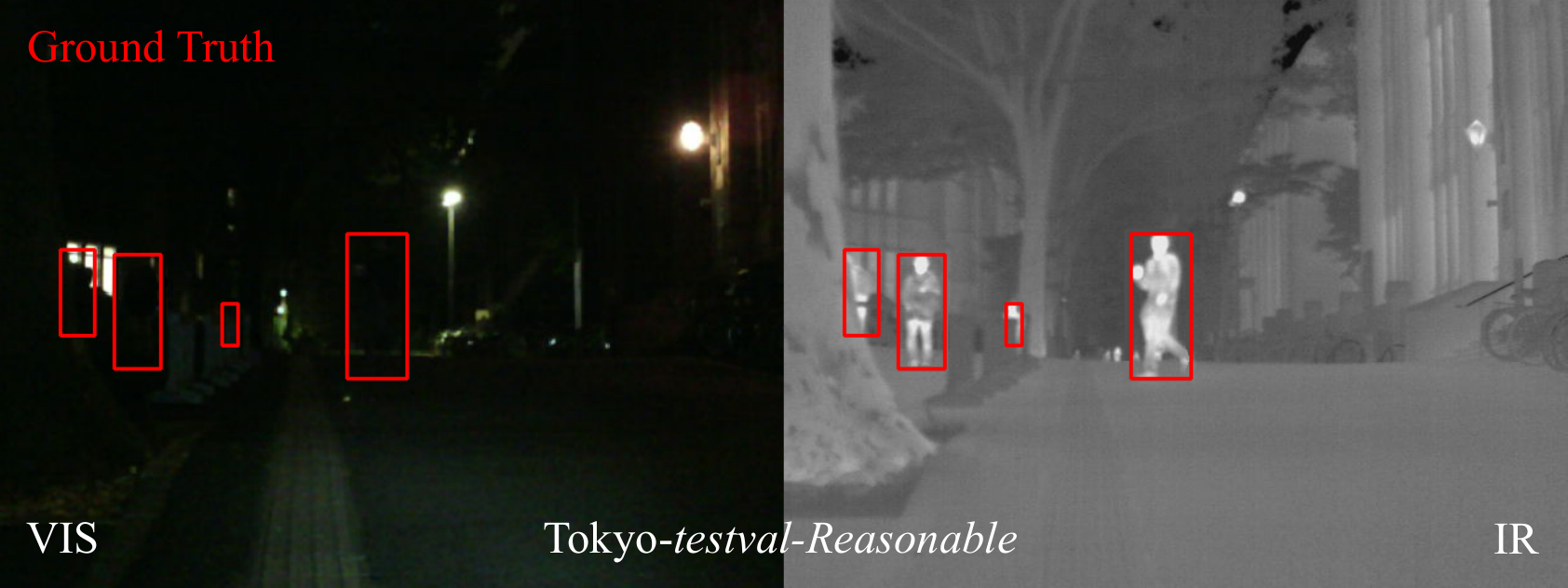}
  \caption{Example image taken from the \mbox{Tokyo-\textit{testval-Reasonable}} dataset with an insufficiently illuminated VIS spectrum causing worse detection performance of the multispectral fusion RPN compared to the pure IR RPN. Detection ground truth is visualized with red bounding boxes.}
					\label{fig:tokyodark}
\end{figure}
\FloatBarrier

\noindent \textbf{Is multispectral fusion always beneficial?} This seems not to be the case. As we can see in Tables~\ref{table:visrpn}, \ref{table:irrpn}, and \ref{table:fusrpn}, the best MR for the Tokyo Multi-spectral Semantic Segmentation dataset is achieved when using the thermal IR images only. While the VIS RPN reaches a MR of 33.62\,\%, the best MR for the IR RPN is significantly lower with 9.04\,\%. This big gap occurs due to a large difference in image quality between the VIS and the IR spectrum: many images acquired at night are insufficiently illuminated in the VIS spectrum. Figure~\ref{fig:tokyodark} shows an example image taken from this dataset. Using such images as input for the multispectral fusion RPN even impairs the results compared to the pure IR RPN as seen in Table~\ref{table:fusrpn}, in which the best MR for the fusion RPN is increased to 10.31\,\%. On the contrary, the images of the KAIST Multispectral Pedestrian Benchmark acquired at nighttime show busy streets with strong illumination by street lights. As a result, the MRs of the pure VIS and IR RPNs are roughly balanced. Improvement for the issue of cross-spectral image quality imbalance could be achieved by following the approach proposed by Li et al.~\cite{Li2018}: the results of VIS, IR, and multispectral fusion RPN are collected individually and combined in a late fusion strategy.

\begin{figure}
\begin{floatrow}
\ffigbox{
  \centering
  \includegraphics[width=.5\textwidth]{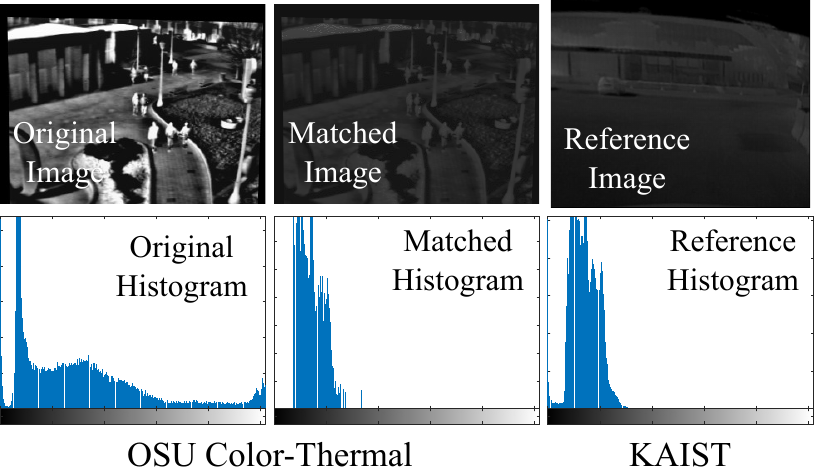}
}{
  \caption{IR histogram matching to narrow the cross-dataset gap between KAIST and OSU dataset.}
	\label{fig:histmatch}
}
\capbtabbox{
  \begin{tabular}{|c|c|c|c|} 
	\hline
	\multicolumn{4}{|c|}{\textbf{KAIST RPN evaluated on OSU dataset}} \\
	\hline
  OSU & VIS & IR & fusion \\
	test dataset & RPN & RPN & RPN \\ 
	\hline
	\hline
  \textit{scale2x-} & \multirow{2}{*}{51.37} & \multirow{2}{*}{80.81} & \multirow{2}{*}{74.13} \\
	\textit{train} & & & \\
	\hline
  \textit{scale2x-test-} & \multirow{2}{*}{11.58} & \multirow{2}{*}{44.88} & \multirow{2}{*}{25.07} \\ 
	\textit{Reasonable} & & & \\
	\hline
	\hline
  \textit{scale2x-hist-} & \multirow{2}{*}{-} & \multirow{2}{*}{69.63} & \multirow{2}{*}{57.07} \\
	\textit{train} & & & \\
	\hline
  \textit{scale2x-hist-} & \multirow{2}{*}{-} & \multirow{2}{*}{22.85} & \multirow{2}{*}{14.45} \\
	\textit{test-Reasonable} & & & \\
	\hline
	\multicolumn{1}{c|}{} & \multicolumn{3}{c|}{\textbf{MR (\%)}}  \\ 
	\cline{2-4}
  \end{tabular}
}{
  \caption{VIS, IR, and fusion RPN trained on KAIST-\textit{train10x} evaluated on OSU dataset.}
	\label{tab:histmatch}
}
\end{floatrow}
\end{figure}
\FloatBarrier

\noindent \textbf{Can we further narrow the gap between datasets?} This is possible with appropriate pre-processing for the thermal IR spectrum. Figure~\ref{fig:histmatch} shows that the IR spectra of the KAIST \emph{Reference Image} and the OSU Color-Thermal \emph{Original Image} have a strongly different appearance. The pixel intensity value distributions in the related histograms verify this observation. Such differences within the IR spectrum seem to be not well handled by the standard pre-processing techniques that are usually applied to VIS images before they are fed as input to a DCNN~\cite{FeiFei2018}. We try to handle this appearance difference by performing histogram matching~\cite{Gonzalez2007}. A reference histogram is generated by averaging sample histograms calculated from  KAIST IR images. This reference histogram is then applied to the IR spectrum of the OSU Color-Thermal images in order to align the appearance with the KAIST IR spectrum as seen in the center column of Fig.~\ref{fig:histmatch}. This approach is indeed beneficial as shown in Table~\ref{tab:histmatch}. We use a VIS, IR, and fusion RPN trained with the KAIST-\textit{train10x} dataset. These RPNs are applied to \mbox{OSU-\textit{scale2x-train}} and \mbox{OSU-\textit{scale2x-test-Reasonable}} achieving 74.13\,\% and 25.07\,\% MR, respectively. We also consider the \mbox{OSU-\textit{scale2x-train}} dataset for the tests as it is by far more challenging compared to \mbox{OSU-\textit{scale2x-test-Reasonable}} as we can clearly see in the MR. Applying the proposed histogram matching reduces the MRs for both the IR RPN and the fusion RPN. The histogram matched OSU Color-Thermal datasets are denoted \mbox{OSU-\textit{scale2x-hist-train}} and \mbox{OSU-\textit{scale2x-hist-test-Reasonable}}. The related MRs for the fusion RPN can be reduced to 57.07\,\% and 14.45\,\%, respectively. This approach is somehow motivated by Herrmann et al.~\cite{Herrmann2018}, who narrowed the KAIST dataset's cross-spectral gap by pre-processing the IR images in a way that a DCNN for person detection in the VIS spectrum needs less training data and epochs for the cross-spectral domain adaptation, i.e. to perform well on the IR images.

\section{CONCLUSIONS}
\label{sec:conclu}

In this paper, we analyzed the generalization ability of RPNs applied to the task of person detection in multispectral images that consist of an aligned color and thermal IR image each. Three publicly available multispectral datasets are chosed. For each, an individual fusion RPN is trained and evaluated across the datasets. The KAIST Multispectral Pedestrian Benchmark dataset proved to be the best dataset to train a well-generalizing fusion RPN. As the fusion RPN needs pure VIS and IR RPNs to be trained before the fusion, we evaluated different pre-training strategies to support the domain adaptation of the RPNs from image classification to person detection. Pre-training is performed for each spectrum individually using popular datasets such as Caltech, CityPerson, or \mbox{CVC-09}. However, there seems to be no benefit in the application of pre-training within the domain of multispectral person detection. Finally, we discussed certain findings showing for example that the quality of the aligned VIS and IR images needs to be balanced in order to benefit from the multispectral fusion and that individual pre-processing of the IR images can improve the fusion results.

\bibliography{report} 
\bibliographystyle{spiebib} 

\end{document}